\documentclass[runningheads]{llncs}
\usepackage{graphicx}
\usepackage{comment}
\usepackage{amsmath,amssymb} 
\usepackage{color}

\usepackage{grffile}
\usepackage{booktabs}
\usepackage{float}

\usepackage[width=122mm,left=12mm,paperwidth=146mm,height=193mm,top=12mm,paperheight=217mm]{geometry}

\newcommand\vocabsize{738}
\newcommand\nparticipants{186}

\begin{document}
\pagestyle{headings}
\mainmatter

\title{Words as Art Materials: \\ Generating Paintings with Sequential GANs} 


%
\institute{SiMiT Lab, Department of Computer Engineering\\Istanbul Technical University, Turkey}


\author{Azmi C. \"{O}zgen,
	Haz{\i}m Kemal Ekenel}

\maketitle

\begin{abstract}
Converting text descriptions into images using Generative Adversarial Networks has become a popular research area. Visually appealing images have been generated successfully in recent years. Inspired by these studies, we investigated the generation of artistic images on a large variance dataset. This dataset includes images with variations, for example, in shape, color, and content. These variations in images provide originality which is an important factor for artistic essence. One major characteristic of our work is that we used keywords as image descriptions, instead of sentences. As the network architecture, we proposed a sequential Generative Adversarial Network model. The first stage of this sequential model processes the word vectors and creates a base image whereas the next stages focus on creating high-resolution artistic-style images without working on word vectors. To deal with the unstable nature of GANs, we proposed a mixture of techniques like Wasserstein loss, spectral normalization, and minibatch discrimination. Ultimately, we were able to generate painting images, which have a variety of styles. We evaluated our results by using the Fréchet Inception Distance score and conducted a user study with \nparticipants{} participants. \footnote{For a demo visit \url{painter-ai.com}}
\keywords{Text-to-Image synthesis, Generative Adversarial Networks (GANs), Sequential GANs, Painting generation}
\end{abstract}

\section{Introduction}

    Generating various data types such as texts, images, audio, videos, and conversion between those have gained popularity with the aid of Generative Adversarial Networks (GANs)~\cite{goodfellow2014generative}. GANs were designed to learn the best representations of given data distribution by training two separate networks called generator and discriminator. The generator is updated to generate similar outputs to given input data while discriminator is updated to distinguish which inputs come from real training data and which come from the generated data. Both networks try to beat the other one and as a result, generator is used to synthesize artificial outputs. The generator might be given a noisy input as well as meaningful data called conditions to convert them into the other desired data type. Converting text (specifically words) to images is the one that we have focused in this work.
    \par
    
\begin{figure}[H]
        \centering
        \includegraphics[width=0.8\linewidth, height=6.5cm]{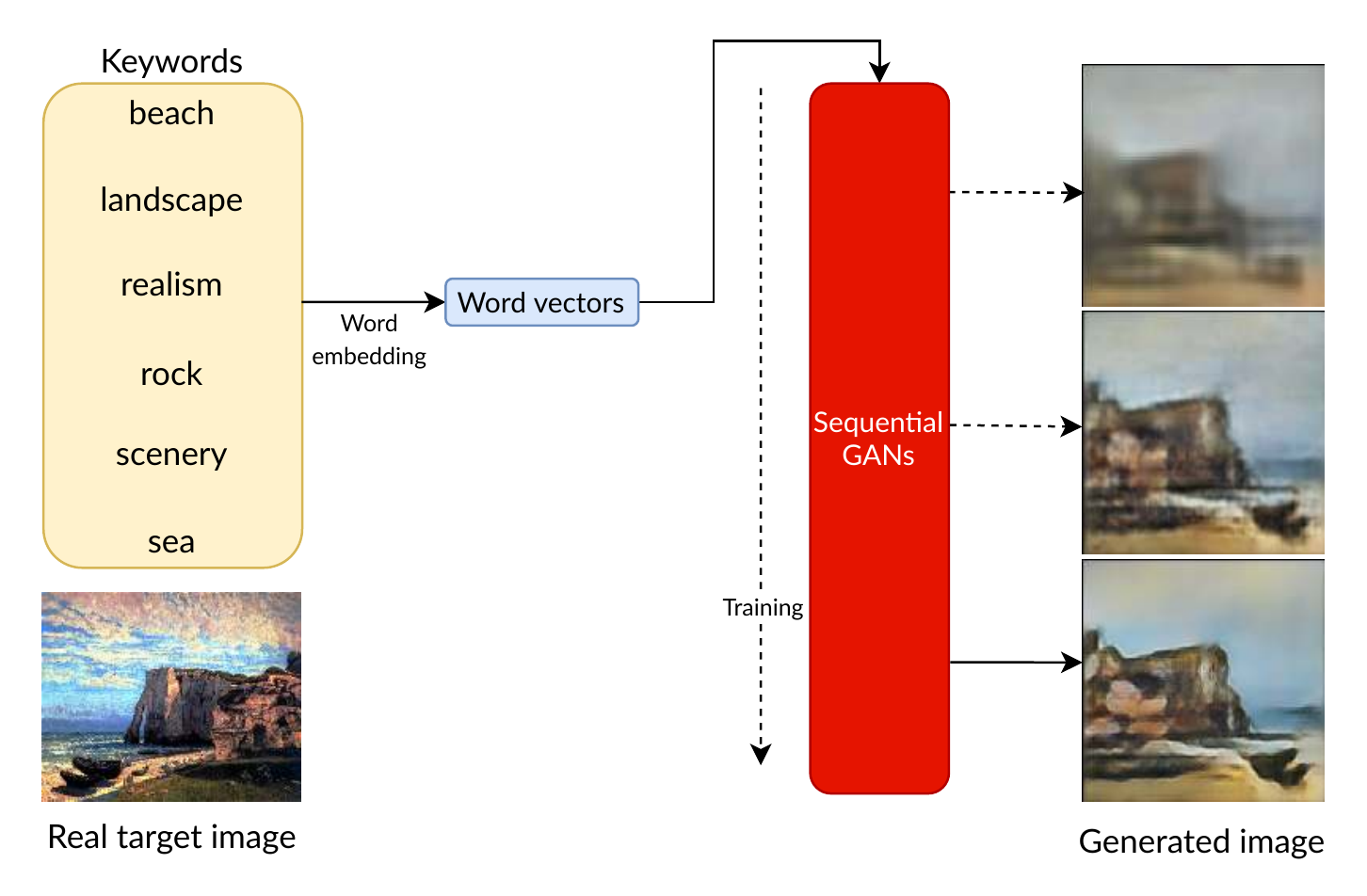}
        \caption{Overview of the proposed pipeline. Keyword lists are converted into word vectors, and they are given to our sequential GAN model. At each stage, model proceeds from the output of the previous stage, and more detailed and higher resolution images are generated. Notice that the generated image contains most of the keywords, and relations between them are creatively connected.}
	    \label{fig:basic_schema}
    \end{figure}

    Image synthesis from given text descriptions is one of the trending research area, which has a considerable amount of published works. Most of these works focused on common datasets, where they try to learn text content and convert them to similar images in the dataset.  On the contrary, we utilized GANs to generate artistic painting images instead of photo-realistic images. Therefore, we have formed our dataset with a large variety of styles in shape, color, drawing technique, and content to have more creativity. Moreover, instead of using descriptive sentences for images, we preferred keywords that reflect the features of the given image. By using only keywords, the word order is insignificant, and relations between them are obscure, so associating the words with the images becomes harder. However, this provides our model the flexibility to generate images more freely which is an important factor for art creation. The overview of the proposed pipeline is shown in Figure \ref{fig:basic_schema}.
    \par
    \par
    
    For the word embedding, we used off-the-shelf Word2Vec~\cite{mikolov2013efficient,mikolov2013distributed} algorithm, whereas, for the generation part, we designed GANs that mostly contain convolutional and deconvolutional layers. In order to learn large variance of images, we proposed sequential architecture of GANs, which consists of three stages. Each stage generates more detailed and higher resolution images as well as adds more artistic styles than the previous one. As the unstable nature of GANs accumulates too fast with such a sequential model, to prevent unbalanced learning curves and the well-known phenomenon called \textit{mode collapse}, we utilized some stabilizing optimization techniques such as Wasserstein loss~\cite{arjovsky2017wasserstein,gulrajani2017improved}, spectral normalization~\cite{miyato2018spectral} and minibatch discrimination~\cite{salimans2016improved}. 

    Our contributions can be summarized as follows: 
    \begin{itemize}
    \item We have created a new keywords-to-painting dataset. Dataset contains $3492$ images with $1746$ digital art images and $1746$ classical painting images. There are $\vocabsize{}$ unique keywords.
    \item We have used keywords for image descriptions, where word order is insignificant and word relations are more obscure unlike full sentence descriptions, leading to more artistic image generations.
    \item We have designed a sequential GAN architecture that handles both word vectors and real-fake image pairs.
    \item We have employed a mixture of optimization techniques ---Wasserstein loss, spectral normalization, and minibatch discrimination--- to train all the networks in a stable manner and efficiently.
    \end{itemize}
    
    The remainder of the paper is organized as follows. In Section \ref{related_works}, we review the related works. Section \ref{proposed_method} is on our proposed pipeline, network architecture, and optimization methods. In Section \ref{experiments}, we present details of the dataset, training techniques, and experimental results. Finally, Section \ref{conclusion} concludes the paper.

\section{Related works}{\label{related_works}}
    \subsubsection{Deep generative learning models} with a convolutional layer structure have shown remarkable performance for generating images. Earlier works were conducted as unsupervised feature learning models~\cite{chen2016infogan,denton2015deep,goodfellow2014generative,liu2017unsupervised,radford2015unsupervised}. Mirza et al.~\cite{mirza2014conditional} introduced a conditional version of this generative model and opened a way for data translation tasks.
    \par
    \subsubsection{Image generation} with labels is extensively studied. Oord et al.~\cite{van2016conditional}, Yan et al.~\cite{yan2016attribute2image} and Odena et al.~\cite{odena2017conditional} have proposed conditional models to generate images where conditions can be class labels or descriptive image attributes. Brock et. al.~\cite{brock2018large} generated class-conditional samples using a large scale GAN training and acquired visually high-quality images. There are also progressive training approaches ~\cite{karnewar2019msg,karras2017progressive}, in which generated output resolutions continually increase and better image qualities are acquired.
    \par
    \subsubsection{Image-to-image translation} studies use images as conditions to generate new images. Isola et al.~\cite{isola2017image} approached this translation as a general-purpose solution and modeled a conditional adversarial network that learns a mapping from input image to output image as well as learns a loss function to train the mapping. Choi et al.~\cite{choi2018stargan} proposed a single GAN model that can perform image-to-image translations for multiple domains. Zhu et al.~\cite{zhu2017unpaired} extended the translation idea for the datasets where the paired input and output samples are absent. Similarly, Yi et al.~\cite{yi2017dualgan} devised an unsupervised general-purpose image-to-image translation network which enables learning for two sets of unlabeled datasets from two different domains.
    \par
    \subsubsection{Style transfer} is another research area that is related to our work since they mostly deal with the translation of artistic essence (colors, styles etc.) of images. \cite{zhu2017unpaired} can also be mentioned here as well because their work was experimented as photo generation from paintings and vice versa. Gatys et al.~\cite{gatys2015neural} designed a CNN-based neural network where they separate and recombine the content and style of images.
    \par
    \subsubsection{Text-to-image synthesis} studies use text descriptions as conditions to GAN models. Dash et al.~\cite{dash2017tac} synthesized images from conditioned text descriptions. Xu et al.~\cite{xu2018attngan} devised an attention-driven mechanism through multi-stage architecture. Similarly, Mansimov et al.~\cite{mansimov2015generating} designed another attentional mechanism with bidirectional recurrent neural networks to encode the image captions. Zhang et al.~\cite{zhang2017stackgan,zhang2018stackgan++} proposed a decomposed architecture where the first stage sketches basic attributes such as color and shape from given text descriptions, and the second stage adds more details and yields better image quality. Qiao et al.~\cite{qiao2019mirrorgan} designed a text-to-image-to-text framework to learn generations by modifying text descriptions. Some other approaches focus on object locations as well as text descriptions while generating images from captions~\cite{hinz2019generating,reed2016learning}. Similarly, Park et al.~\cite{park2018mc}  proposed a model that focuses on foreground and background objects at the same time for given text descriptions. For better learning of text descriptions, using mismatching text is another idea~\cite{reed2016generative}.

\section{Proposed Method}
\begin{figure}[ht]
            \centering
            \includegraphics[width=0.70\linewidth, height=7cm]{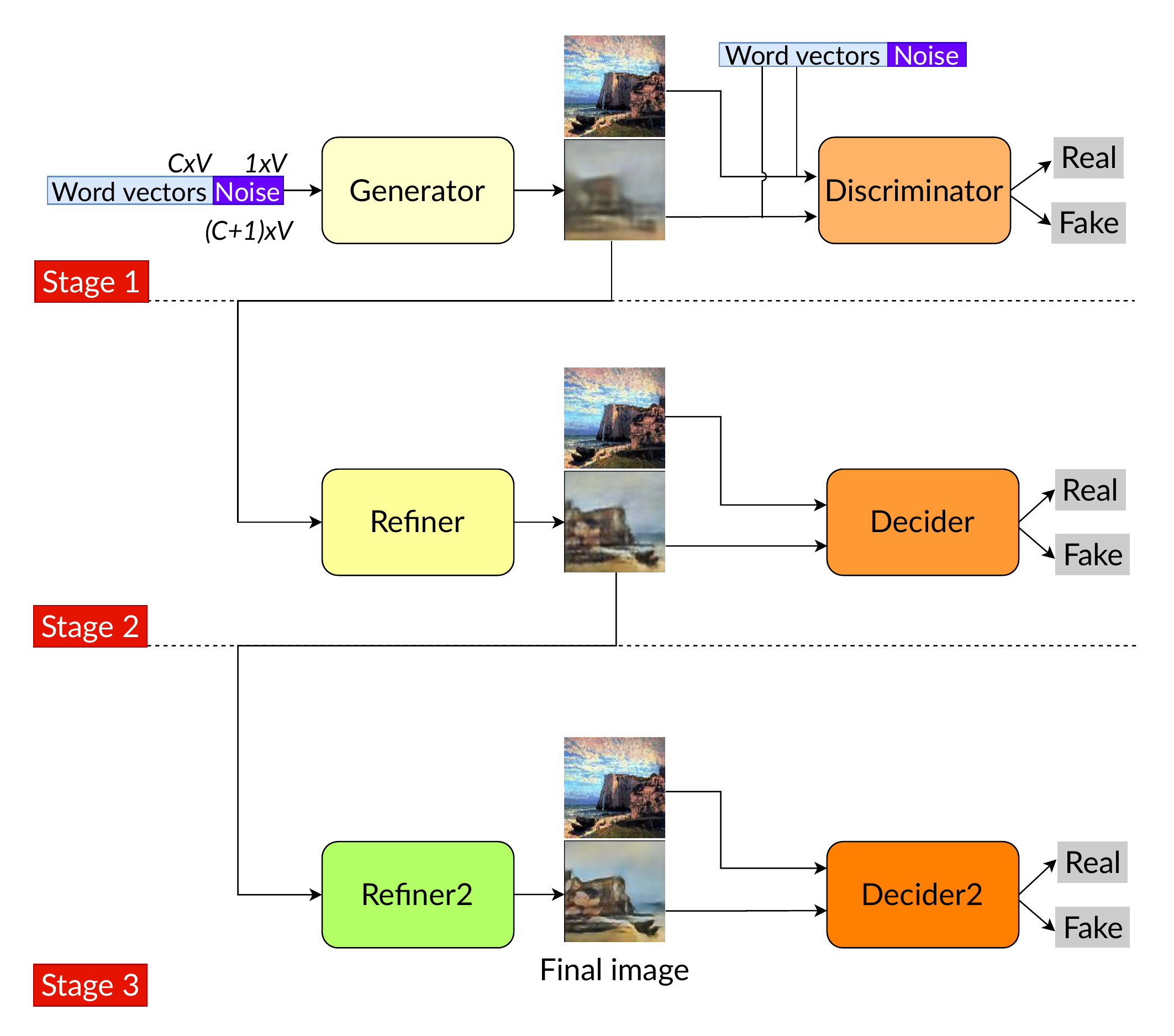}
            \caption{Basic model workflow with three stages of GANs. Discriminator outputs are not bounded in an interval because of Wasserstein loss (Real-Fake separation is acquired by a sigmoid function that is not internally in networks.)}
    	    \label{fig:model_workflow}
        \end{figure}

{\label{proposed_method}}
    Our method consists of two main parts: Word embedding and three stages of GANs. The proposed method's workflow is illustrated in
    Figure \ref{fig:model_workflow}, and it is explained in the following subsections.

    \subsection{Word Embedding}
        A shallow neural network algorithm Word2Vec~\cite{mikolov2013efficient,mikolov2013distributed} was used for word embedding. The basic architecture is a two-layer neural network with $V$ dimensional output layer. $V$ also represents the size of word vectors, which is taken as $64$. Context size (number of words representing one image) is fixed to six, and one noise vector is concatenated to the end of context as the same size of a real word vector size. Finally, our vocabulary size is $\vocabsize$ (total unique words) for the training dataset.
        
    \subsection{Model Architecture}
        The proposed model architecture consists of three staged GANs. The first stage processes the word vectors in the generator and produces a relatively lower resolution image. This image forms a base for the upcoming stages. On the other hand, discriminator processes real image-word vectors and fake image-word vectors pairs and gives unbounded outputs. Unbounded output without sigmoid function at the end is used since we use Wasserstein loss as we will discuss in Section~\ref{model_optimization}. The second stage takes the output of the first stage generator, refines it, and outputs a higher resolution image. This stage's networks are called as refiner and decider even though they are generator and discriminator essentially. At this stage, no word vector is processed by neither refiner nor decider. The third stage is quite similar to the second stage. It takes the output of the second stage and gives the final image with the highest resolution. The advantage of using this type of sequential staged model is that separating base image creation with word vectors and image refinement processes. Therefore, we can design more specialized networks and increasing image resolution through the layers becomes easier. Model architecture is similar to the one proposed by Zhang et al.~\cite{zhang2017stackgan,zhang2018stackgan++} with one extra refiner stage, and there are some differences at the inner structure of GANs and optimization techniques as we will discuss in the upcoming sections.

        \subsubsection{The First Stage} 
\begin{figure}[ht]
            \centering
            \includegraphics[width=0.90\linewidth, height=5cm]{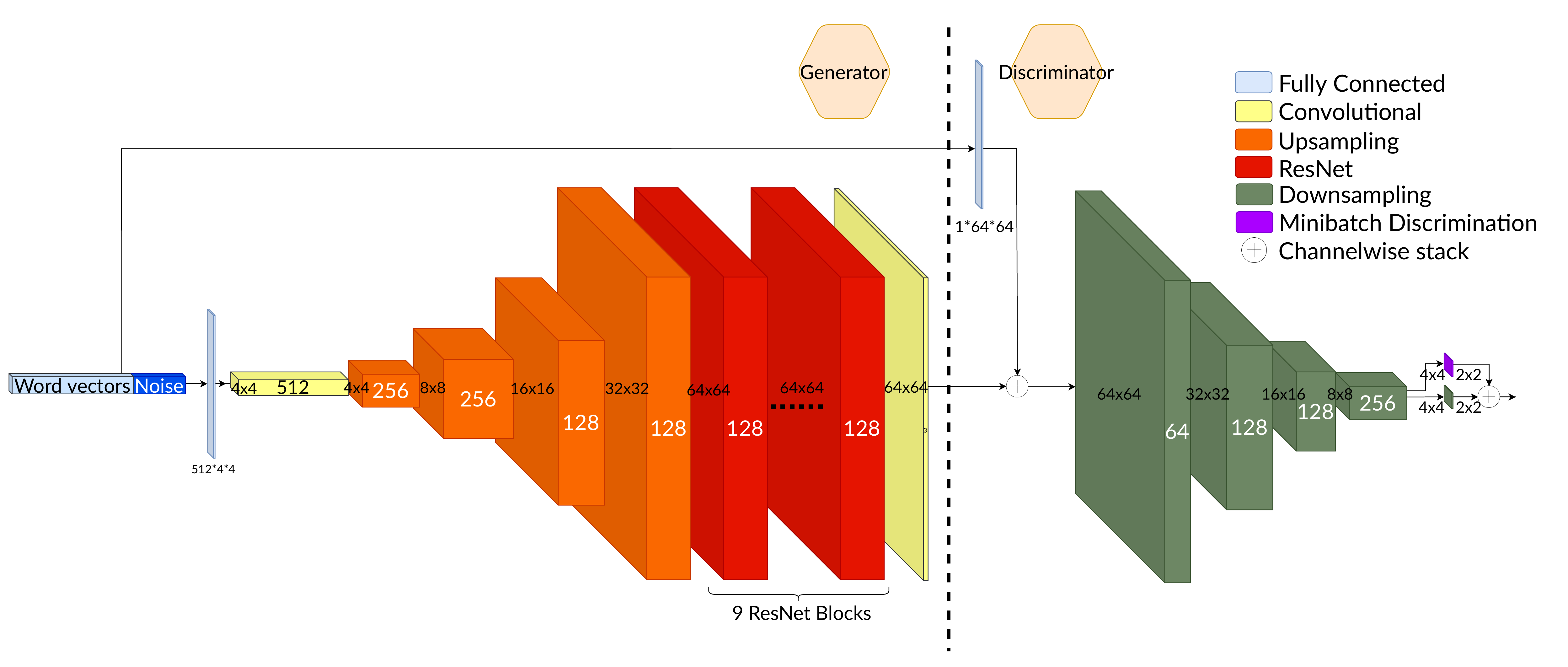}
            \caption{The first stage GAN architecture. Generator has one fully connected layer to process word vectors, convolutional upsample layers and ResNet blocks. Discriminator has one fully connected layer to process word vectors, convolutional downsample blocks and minibatch discrimination layer. Batch normalization is applied after convolutional layers except the last layer and the first layer of the generator and discriminator.}
    	    \label{fig:first_stage_GAN}
        \end{figure}

architecture is given in Figure \ref{fig:first_stage_GAN}. Word vector inputs are fixed in the shape of $7 \times 64$ with six words and one additional noise row. If given words for the image is not filling the context of six words, then one of the closest words from vocabulary is chosen randomly. On the contrary, if there are more words than six, then a random subset is chosen. This way inputs become noisier and varied so that it provides stable learning curves as proposed by Salimans et al.~\cite{salimans2016improved}. The generator outputs a $64 \times 64$ image. The same word vectors are given to discriminator paired with real and fake images separately. 

        \subsubsection{The Second Stage} 
\begin{figure}[ht]
            \centering
            \includegraphics[width=0.90\linewidth, height=4.3cm]{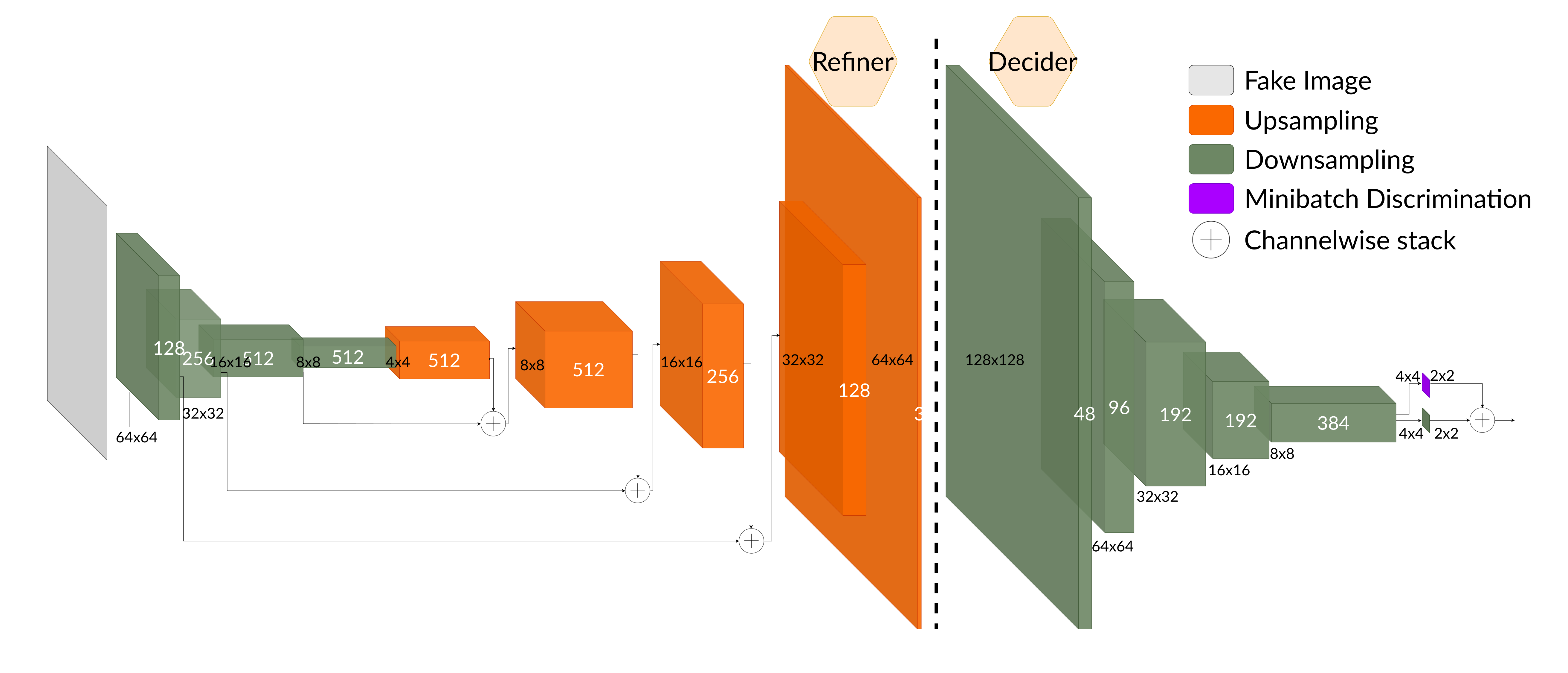}
            \caption{The second stage GAN architecture. Generator has convolutional upsample and downsample blocks. There are skip connections between layers that has same resolution images. Discriminator has convolutional downsample blocks and minibatch discrimination layer. Batch normalization is applied after convolutional layers except the last layer and the first layer of the generator and discriminator.}
    	    \label{fig:second_stage_GAN}
        \end{figure}

architecture refines the first stage outputs and gives a $128 \times 128$ image. Unlike the first stage, it has skip connections in the generator part. This provides smooth gradient flow between these layers and faster learning. Since this stage proceeds where the first stage left off, word vectors are not processed at this stage but only the output of the first stage. Detailed schema is illustrated in Figure \ref{fig:second_stage_GAN}.
            
        \subsubsection{The Third Stage} architecture is almost identical to the second stage as given in Figure \ref{fig:second_stage_GAN} except that it is fed with the output of the second stage and generates a $256 \times 256$ image.
        
    \subsection{Model Optimization} \label{model_optimization}
        Stable training of GANs is challenging since it is highly dependent on parameters. When either one of the generator or discriminator starts to dominate training, \textit{mode collapse} is quite likely to occur. Our architecture has three separate GANs, and the first two feed the others by their outputs. Therefore, some stabilizing optimization techniques are necessary. 
        
        \subsubsection{Wasserstein loss with gradient penalty (WGAN-GP)}~\cite{arjovsky2017wasserstein,gulrajani2017improved} were selected for optimization. As proposed by Goodfellow et al.~\cite{mirza2014conditional}, classical GAN loss is given as in Equation \ref{minimax_loss}:
        
        \begin{equation}\label{minimax_loss}
            \mathbb{E}_x[\log(D(x))] + \mathbb{E}_z[\log(1 - D(G(z)))]
        \end{equation}
        
        where $x$ and $z$ represent the real and noise inputs, respectively. The generator tries to minimize and the discriminator tries to maximize this, whereas, in Wasserstein loss the discriminator tries to maximize $D(x) - D(G(z))$ and the generator tries to maximize $D(G(z))$. As we do not use sigmoid at the output of any discriminator network, the loss is not bounded. Therefore, as proposed by Gulrajani et al.~\cite{gulrajani2017improved} we apply the gradient penalty as an addition to loss. The final form of the loss function is given in Equation \ref{wasserstein_loss}.
        
        \begin{equation}\label{wasserstein_loss}
            \mathbb{E}_{\tilde{x}}[D(\tilde{x})] - \mathbb{E}_x[D(x)] + \lambda\mathbb{E}_{\tilde{x}}[(||\nabla_{\tilde{x}}D(\tilde{x})||_2 - 1)^2]
        \end{equation}

        where $x$ and $\tilde{x}$ represent real and fake data instances respectively. $\lambda$ is constant multiplier of gradient penalty and taken as $10.0$. Wasserstein loss provided us more stable converging curves as opposed to BCE (Binary Cross Entropy) loss and LSGAN loss~\cite{mao2017least}.
        \par
        \subsubsection{Minibatch discrimination}~\cite{salimans2016improved} is the other technique related to the architectures shown in Figures \ref{fig:first_stage_GAN} and \ref{fig:second_stage_GAN}, which forces the generator to generate a wide variety of images and prevent mode collapse. It is applied to final dense layer of the discriminator by means of concatenating the similarity of input image $x$ with all other images in the batch. Similarity for the minibatch $b$ is given as
        
        \begin{equation}\label{minibatch_similarity}
            o\left ( \boldsymbol{x}_i \right )_b = \sum_{j=1}^n c_b\left ( \boldsymbol{x}_i, \boldsymbol{x_j} \right ) \in \mathbb{R}
        \end{equation}
        
        where $c_b\left ( \boldsymbol{x_i}, \boldsymbol{x_j} \right ) = \exp\left ( -\left \| \boldsymbol{M}_{i,b} - \boldsymbol{M}_{j,b} \right \|_{L_1} \right ) \in \mathbb{R}$ is $L_1$-distance between $i^{th}$ and $j^{th}$ image. Here, $\boldsymbol{f\left ( \boldsymbol{x_i} \right )} \in \mathbb{R}^A$ is the vector of features of input $\boldsymbol{x}_i$, $\boldsymbol{T} \in \mathbb{R}^{A \times B \times C}$ is learnable parameter tensor of minibatch discrimination layer and $ \boldsymbol{M}_i = \boldsymbol{f}_i \cdot \boldsymbol{T} $. Similarity output of the all batches is given as
        
        \begin{equation}\label{batch_similarity}
            o\left ( \boldsymbol{x_i} \right ) = \left [ o\left ( \boldsymbol{x_i} \right )_1, o\left ( \boldsymbol{x_i} \right )_2, \dots, o\left ( \boldsymbol{x_i} \right )_B \right ] \in \mathbb{R}^B
        \end{equation}
        
        which is concatenated channel-wise with the feature vector $\boldsymbol{f\left ( \boldsymbol{x_i} \right )}$.\\
        As our dataset contains a wide variety of images, real inputs to the discriminator will have very low similarity. On the contrary, if the generator starts to output monotonic images, the similarity metric given in Equation \ref{batch_similarity} will be very high. In this case, discriminator should penalize the generator. As we shall see in Section~\ref{experiments}, the outputs of our generators are very unique.
        \par
        \subsubsection{Spectral normalization}~\cite{miyato2018spectral} was applied to all convolutional layers as an additional stabilizer of training. As Miyato et al. stated spectral normalization provides more diversely generated samples than the conventional weight normalization~\cite{miyato2018spectral}. They compared their model Spectral Normalization GANs (SN-GANs)~\cite{miyato2018spectral} to WGAN-GP models and regularization techniques and proposed that the images generated by their models are clearer and more diverse on CIFAR-10~\cite{krizhevsky2009learning} and STL-10~\cite{coates2011analysis} datasets. In this work, we utilized spectral normalization in combination with the gradient penalty as a regularization method.
        \par
        \subsubsection{Separate parameters} for the generator and the discriminator is another technique to have more stable training. As Heusel et al. introduced~\cite{heusel2017gans} using different learning rates (generator: $0.0001$, discriminator: $0.0002$) solves the slow learning curves of the discriminator as opposed to the generator. Inspired by this idea, we applied different dropout rates (generator: $0.2$, discriminator: $0.65$) for all convolutional and dense layers on all networks. Finally, we flipped labels with $0.05$ probability (fake images become real and real ones become fake) for again stabilizing training and preventing mode collapse. 

\section{Experiments}{\label{experiments}}

\begin{figure}[ht]
        \centering
        \includegraphics[width=0.80\linewidth, height=3.7cm]{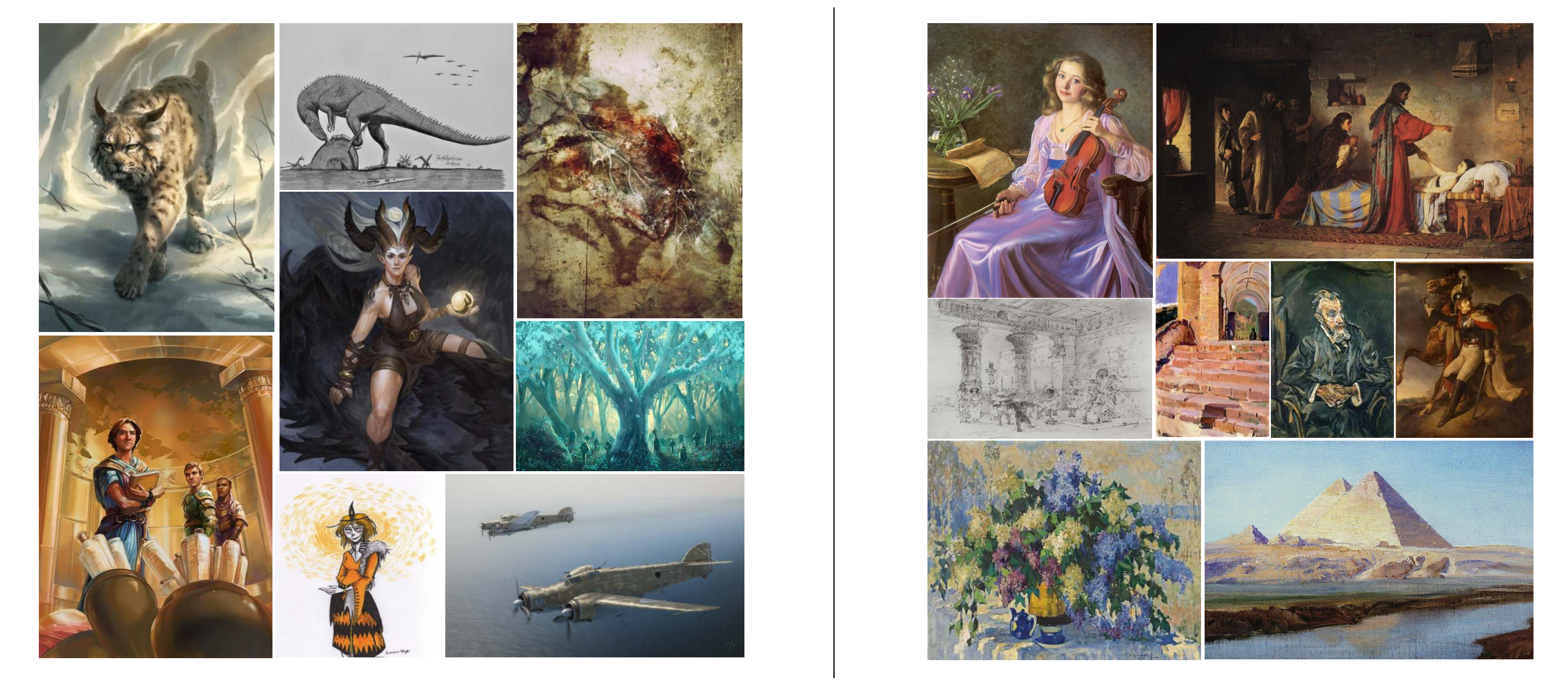}
        \caption{Sample digital art images on the left and classical art images on the right}
	    \label{fig:dataset}
    \end{figure}

\begin{figure}[ht]
        \centering
        \includegraphics[width=0.90\linewidth, height=6.7cm]{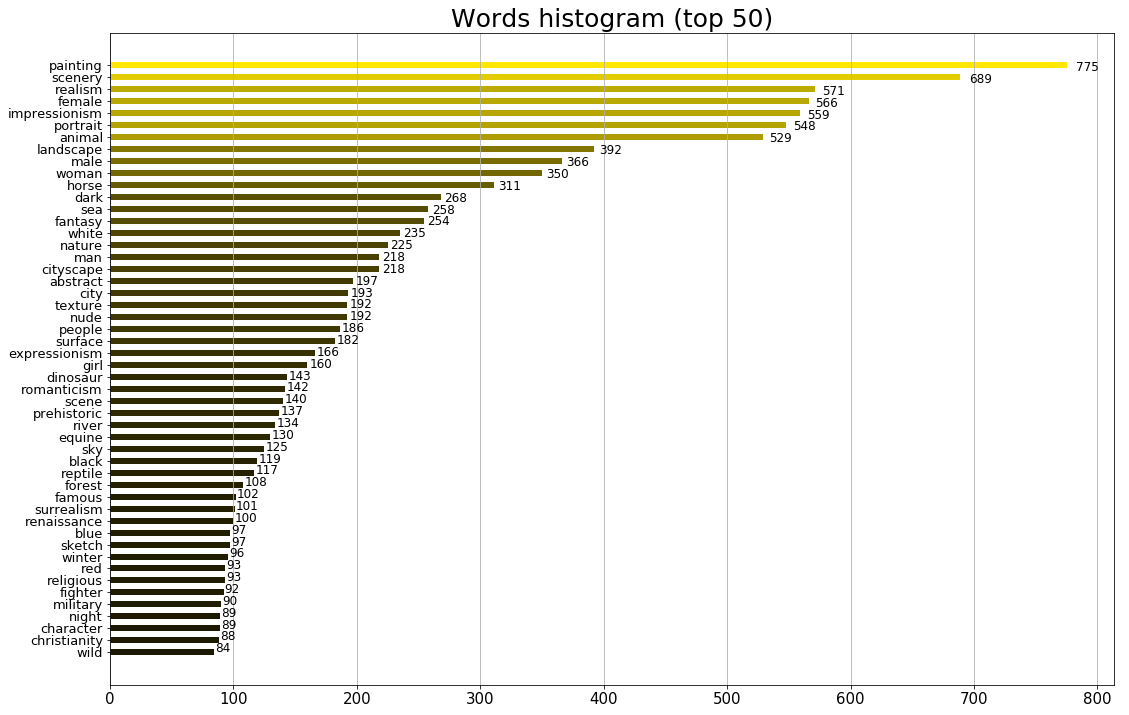}
        \caption{Top 50 the most used words in our dataset.}
	    \label{fig:words_hist}
    \end{figure}

Our dataset\footnote{Accessible from \url{https://doi.org/10.5281/zenodo.3690752}} is a painting images collection acquired from \textit{deviantart.com} for digital paintings and \textit{wikiart.org} for classical paintings. $1746$ digital paintings and $1746$ classical paintings were used in total and $349$ images were reserved for testing. As can be seen in Figure \ref{fig:dataset}, our dataset contains a large variety of images in shape, color, content, and style. Similarly, our vocabulary size ($\vocabsize$) is quite extensive. Words histogram of our dataset is shown in Figure~\ref{fig:words_hist}. We believe that this variety is a necessity for artistic essence since it leads to generating unique images for a given set of input words.
    
    \subsubsection{Data augmentation} plays an important role in our experiments due to the relatively small size of our dataset. We applied Gaussian blur with a randomly selected radius from the uniformly distributed $(0, 3)$ interval. Color jitter was used that randomly factored brightness, contrast, saturation, and hue values in the intervals $(0.6, 1.6)$, $(0.6, 1.6)$, $(0.6, 1.6)$, and $(-0.2, 0.2)$ respectively. We did not use other well-known augmentation methods, such as horizontal flipping, image rotation, channel swapping or grayscaling since they would have caused incompatibility between words and content of the images. Images are normalized to $(0, 1)$ scale as well as our word vectors.
    \par
    \subsubsection{Training} of all three stages was conducted simultaneously. The alternating training method was not used. Namely, all the generators and the discriminators were updated at the end of the same minibatch. Instead, we applied separation of learning rate and dropout rate parameters to organize stabilized training. For all networks, Adam~\cite{kingma2014adam} optimizer with $beta=0.5$ was used. The batch size is $16$, and all networks were trained for $3000$ epochs. Learning rates were initialized to $0.0001$ and $0.0002$ for the generator and the discriminator, respectively, and decayed to half for every $300$ epochs but not below $0.00001$.
    \par
    \subsubsection{Fréchet Inception Distance (FID)}~\cite{heusel2017gans} is the first method that we used to measure the quality of the generated images. Lower FID score corresponds to more similar real and generated samples as measured by the Inception-v3 model~\cite{szegedy2016rethinking}. It evaluates the similarity between the real and the generated datasets by measuring the distance of two Gaussian distribution which is given as
    
    \begin{equation}\label{FID}
            ||\mu_r - \mu_g||^2 + Tr(C_r + C_g - 2(C_{r}C_g)^{1/2})
    \end{equation}
    
    where $(\mu_r, C_r)$, $(\mu_g, C_g)$ are the mean and covariance of the real and generated image distributions. We compared the quality of outputs by FID score (lower is better) in Table \ref{tab:fid}. As it can be seen from Table \ref{tab:fid}, more realistic images are generated through the stages of our model.

    \begin{table}[H]
        \centering
        \begin{tabular}{@{}cc@{}}
            \toprule
                             & FID score       \\ \midrule
            Random noise     & 487.66          \\
            Stage 1          & 384.67          \\
            Stage 2          & 234.18          \\
            \textbf{Stage 3 (Final output)} & \textbf{218.34} \\ \bottomrule
        \end{tabular}
        \caption{FID scores of random noise and stages of our model.}
        \label{tab:fid}
    \end{table}
    
    \subsubsection{Survey}
\begin{figure}[ht]
        \centering
        \includegraphics[width=0.80\linewidth, height=4.4cm]{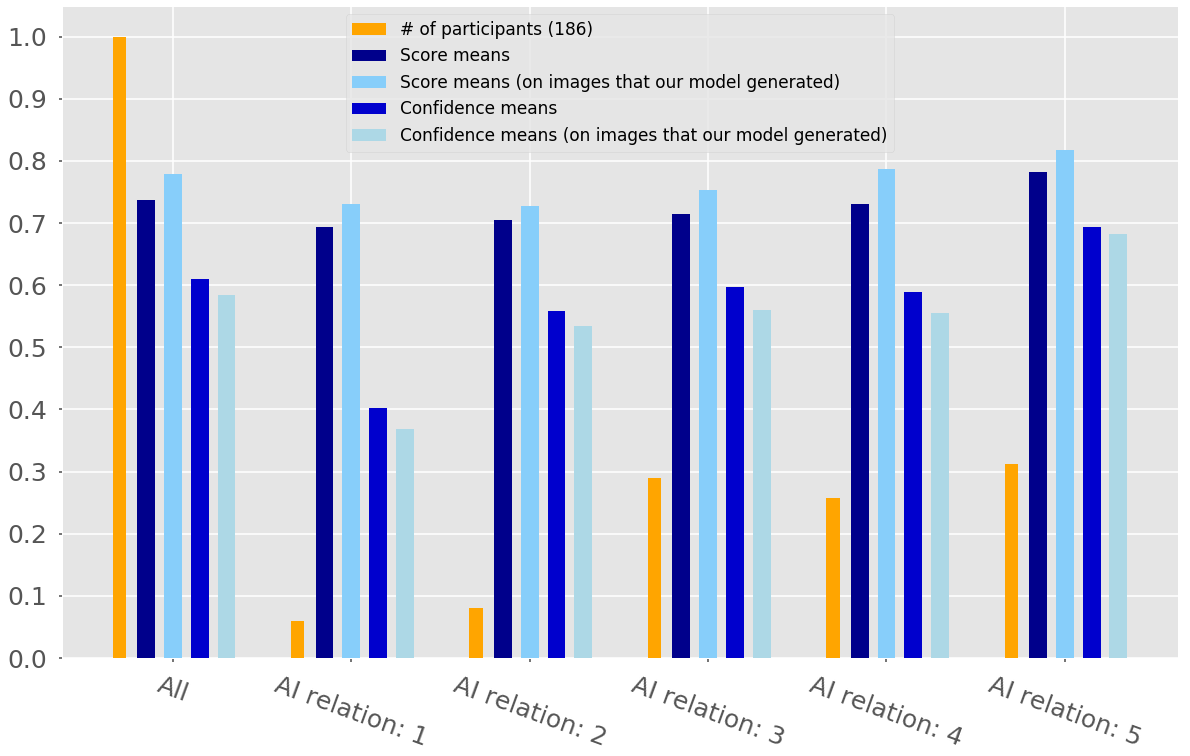}
        \caption{Survey results. From left to right; all participants and groups according to their AI relations from 1 to 5. Blue bars represent the normalized total score of \nparticipants{} participants that they acquire from 48 images. Light blue bars represent total score on generated 24 images. All bars are normalized to $(0,1)$ interval.}
	    \label{fig:survey_ai}
    \end{figure}

\footnote{Accessible from \url{https://forms.gle/LiWTiMKbLvsx6ZL38}} was conducted as a second method to measure the quality of generated images. The survey was held by querying the participants whether a given painting image is made by our model or by a human. Besides, we asked their confidence level on each image (\textit{Unsure}, \textit{Somewhat confident}, \textit{Very confident}). Input words of the images were not given in the survey. Results were collected from \nparticipants{} participants where $98\%$ of the total is at least a university student or has a college degree. Participants were shown 48 images. 24 of them are generated by our model and the rest is made by humans (those images were collected from our dataset but were not used in the training). The participants were asked their relations with AI at a  scale of 1 to 5. Survey results according to participant relations with AI can be seen in Figure \ref{fig:survey_ai}.
    
    \begin{figure}[ht]
        \centering
        \includegraphics[width=0.80\linewidth, height=4.4cm]{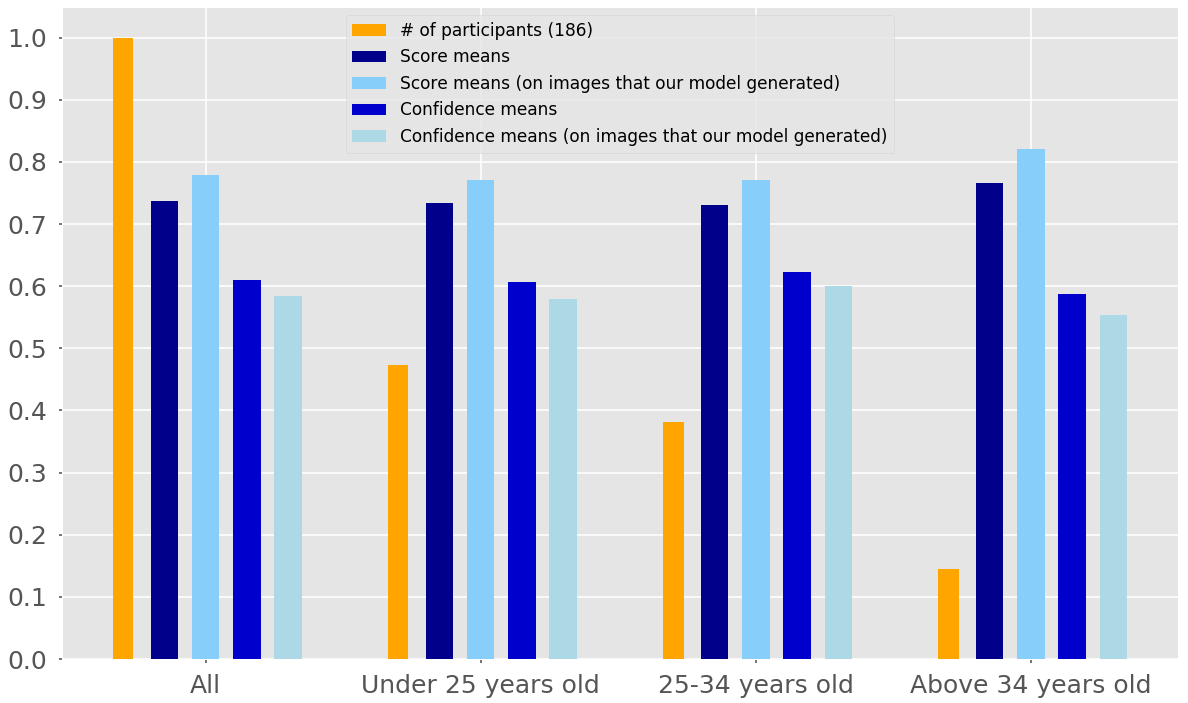}
        \caption{Survey results. From left to right; all participants and groups according to their age groups. Blue bars represent the normalized total score of \nparticipants{} participants that they acquire from 48 images. Light blue bars represent total score on generated 24 images. All bars are normalized to $(0,1)$ interval.}
	    \label{fig:survey_age}
    \end{figure}
    
\begin{figure}[H]
        \centering
        \includegraphics[width=0.75\linewidth, height=2.1cm]{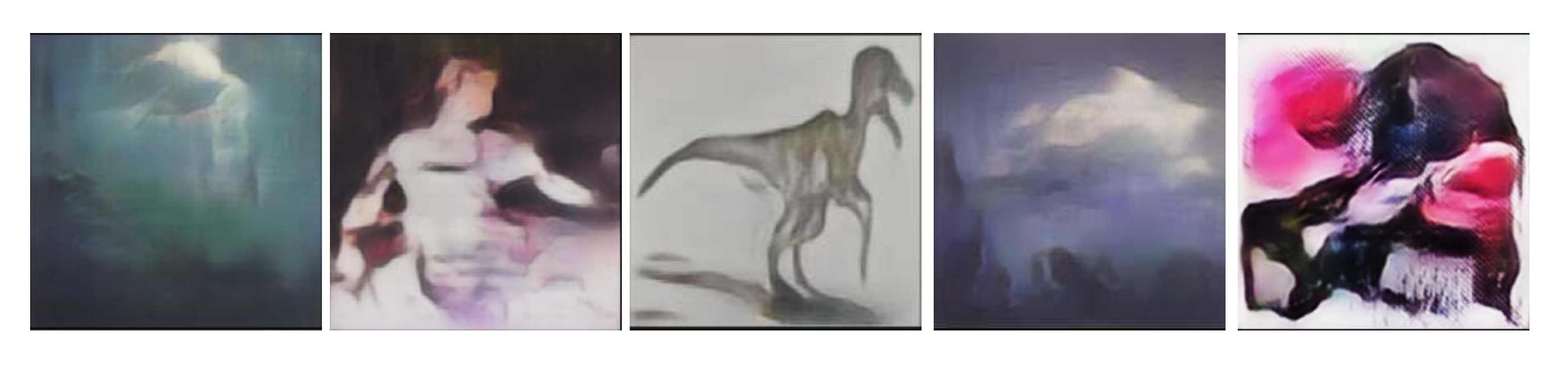}
        \caption{Top 5 the most confused images that are actually made by our model. From left to right; $46.2\%$, $40.3\%$, $39.2\%$, $38.2\%$ and $30.6\%$ of all participants guessed that they are made by human. We are informed that most of the participants thought that these images have quite realistic brush strokes and objects are not obscure as the others.}
	    \label{fig:best_ai}
    \end{figure}

\begin{figure}[H]
        \centering
        \includegraphics[width=0.75\linewidth, height=2.1cm]{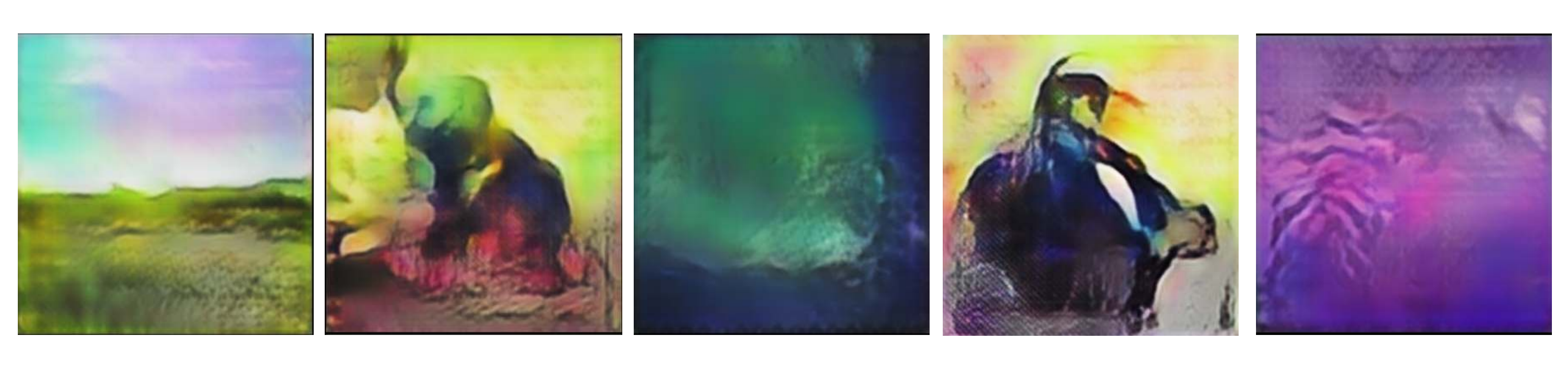}
        \caption{Top 5 the least confused images that are actually made by our model. From left to right; $6.5\%$, $10.8\%$, $11.3\%$, $11.8\%$ and $12.4\%$ of all participants guessed that they are made by human. For those images, participants said that the artifacts around the images played an important role in their decisions.}
	    \label{fig:worst_ai}
    \end{figure}

\begin{figure}[H]
        \centering
        \includegraphics[width=0.75\linewidth, height=2.1cm]{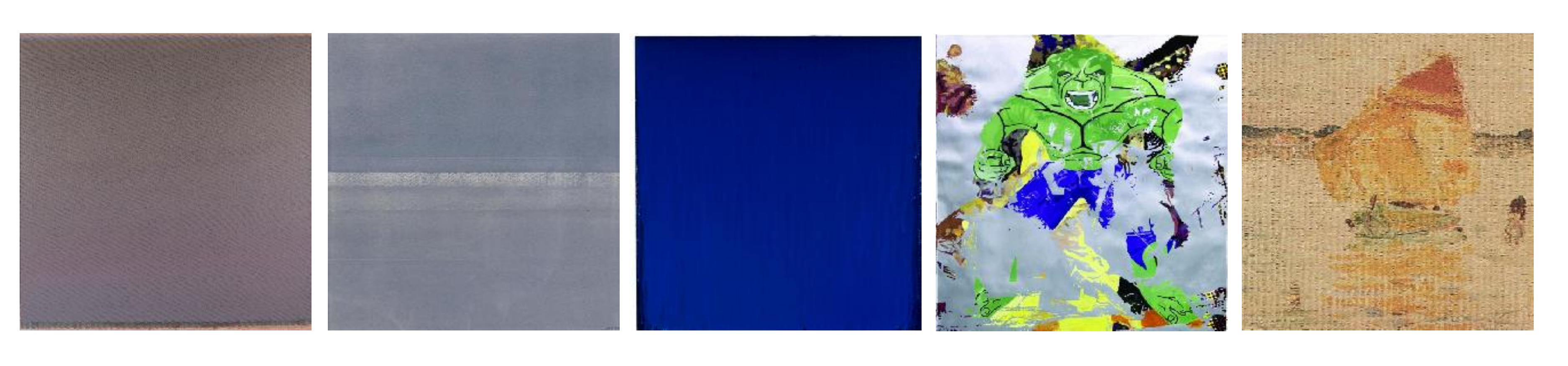}
        \caption{Top 5 the most confused images that are actually made by humans. From left to right; $76.9\%$, $59.1\%$, $58.6\%$, $58.1\%$ and $51.1\%$ of all participants guessed that they are made by our model. Plain, monochromatic and worn images make participants suppose that there must be an artificial generation.}
	    \label{fig:best_human}
    \end{figure}

\begin{figure}[H]
        \centering
        \includegraphics[width=0.75\linewidth, height=2.1cm]{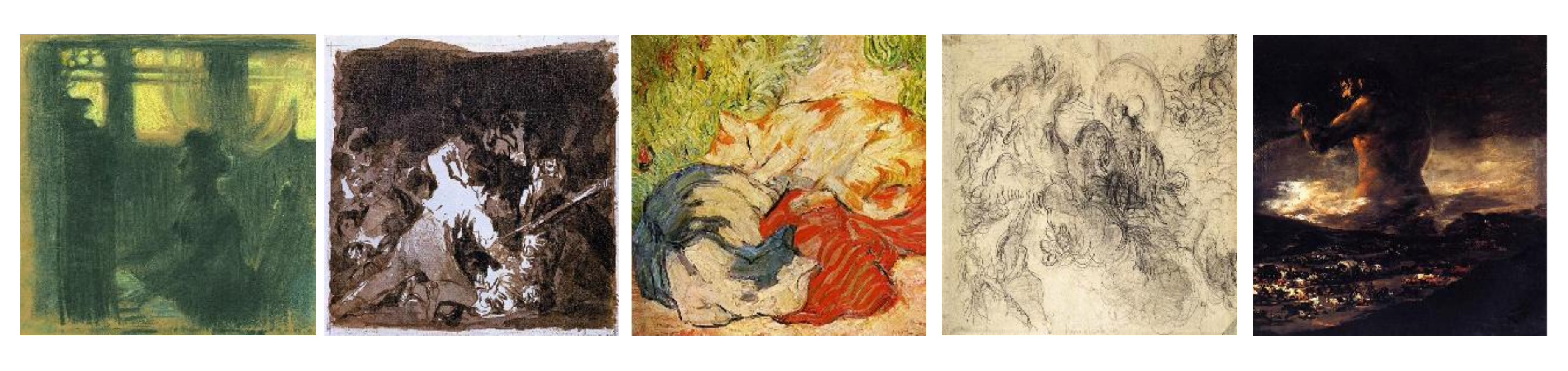}
        \caption{Top 5 the least confused images that are actually made by humans. From left to right; $8.6\%$, $10.8\%$, $10.8\%$, $14.0\%$ and $14.5\%$ of all participants guessed that they are made by our model. Color usage on the canvas and theme of the paintings is very critical in the thought process of all the images. Especially, if the participant has experience with artificially generated images before or has an advanced understanding of art, it is more difficult to deceive them. Most of the participants said that these images are more artistic and have more details compared to others.}
	    \label{fig:worst_human}
    \end{figure}

    As a result of the survey, the average success rate of guessing whether a given image (of all 48 images) is generated by our model or by a human is found to be $73.8\%$. The success rate of guessing correctly for all images (24 images) generated by our model is $77.9\%$. According to survey results, the top five the most and the least confused images created by both our model and humans are given in Figures \ref{fig:best_ai}, \ref{fig:worst_ai}, \ref{fig:best_human}, and \ref{fig:worst_human}, respectively.
    
\begin{figure}[ht]
        \centering
        \includegraphics[width=0.99\linewidth, height=15.0cm]{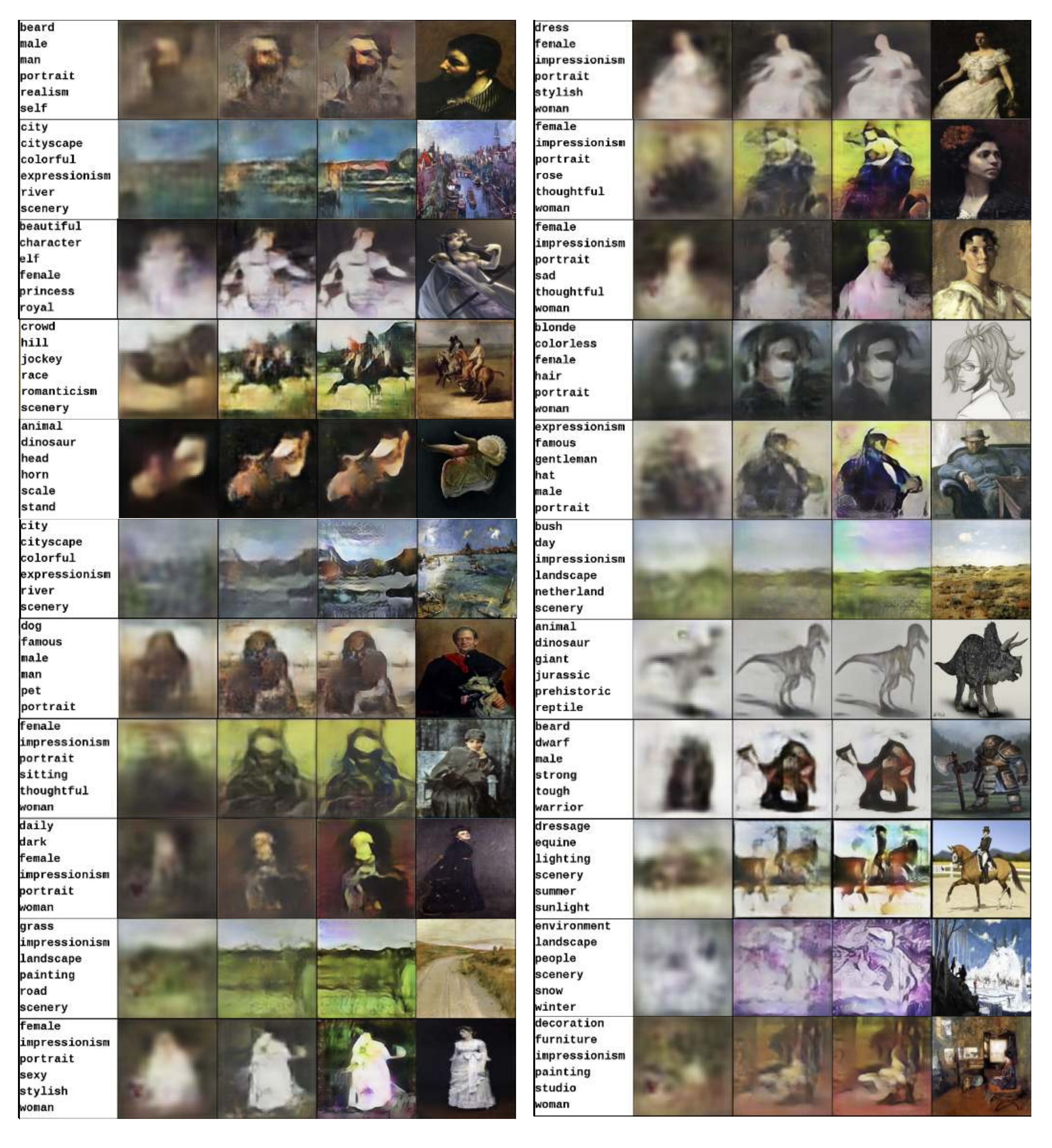}
        \caption{Sample results. From left to right (as two separate columns); given words, our outputs through three stages and corresponding ground truth for given words. The image content of the generated images is compatible with given words generally. Besides, when a very similar word set is given (on the right column, top four paintings) unique styles and contents were generated. As a weakness, some details like the face of people are not created very well and there are artifacts around the objects. We believe that those weaknesses would be resolved by expanding the dataset.}
	    \label{fig:results}
    \end{figure}

    From the visual results of the survey, we can comment that bright colors make participants think that image is artificially generated. Besides, when the objects are prominently seen and artifacts around the images are absent, participants can be deceived. More detailed visual outputs for given words at different stages are provided in Figure \ref{fig:results}.


\nopagebreak
\section{Conclusion}{\label{conclusion}}
    We proposed a sequential GAN model to generate paintings from keywords. For this purpose, we created a keyword-to-painting dataset. Our model acquired $218.34$ FID score, and at the survey with 48 images (24 images were made by our model, and the rest were made by humans) $26.2\%$ of all 186 participants mistaken about who made it. Currently, our model capability should be evaluated by its visual outputs even though we have provided an FID score for quantitative measurement and a survey. Please note that we cannot compare the FID score to the other works in literature due to the use of different datasets. However, we believe visual outputs show promising results as well. Please note that with larger datasets, variety in inputs would increase, and this would in return lead to producing better images. Nevertheless, visual results show that our model is quite successful at generating visually appealing, mixed style, fairly suited to text descriptions painting images.
    
\clearpage
%
%
\bibliographystyle{splncs04}
\bibliography{main}
\end{document}